\documentclass[letterpaper]{article} 
\usepackage{aaai2026}  
\usepackage{times}  
\usepackage{helvet}  
\usepackage{courier}  
\usepackage[hyphens]{url}  
\usepackage{graphicx} 
\usepackage{natbib}  
\usepackage{caption} 
\usepackage{algorithm}
\usepackage{algorithmic}

\usepackage{amsthm,amsmath,amssymb}
\usepackage{cite}
\usepackage{booktabs}
\usepackage{multirow}
\usepackage{subfigure}
\usepackage[misc]{ifsym}
\usepackage{xcolor}
\usepackage{pifont}

\usepackage{newfloat}
\usepackage{listings}
\DeclareCaptionStyle{ruled}{labelfont=normalfont,labelsep=colon,strut=off} 
\floatstyle{ruled}
\newfloat{listing}{tb}{lst}{}
\floatname{listing}{Listing}
\title{Semi-Supervised Synthetic Data Generation with Fine-Grained Relevance Control for Short Video Search Relevance Modeling}
\author{
    Haoran Li\textsuperscript{\rm 1,2},
    Zhiming Su\textsuperscript{\rm 2},
    Junyan Yao\textsuperscript{\rm 2},
    Enwei Zhang\textsuperscript{\rm 2},
    Yang Ji\textsuperscript{\rm 2},
    Yan Chen\textsuperscript{\rm 2},
    Kan Zhou\textsuperscript{\rm 2},
    Chao Feng\textsuperscript{\rm 2}\thanks{Corresponding author.},
    Jiao Ran\textsuperscript{\rm 2}
}
\affiliations{
    \textsuperscript{\rm 1}National Key Laboratory for Multimedia Information Processing, School of Computer Science, Peking University\\
    \textsuperscript{\rm 2}ByteDance Douyin Content Group \\
    haoranli@stu.pku.edu.cn, \{sunzhenming, yaojunyan.spacefish, enweizhang, chensong.09, chaofeng.zz, ranjiao\} @bytedance.com

}

\usepackage{bibentry}

\begin{document}

\maketitle

\begin{abstract}
Synthetic data is widely adopted in embedding models to ensure diversity in training data distributions across dimensions such as difficulty, length, and language. However, existing prompt-based synthesis methods struggle to capture domain-specific data distributions, particularly in data-scarce domains, and often overlook fine-grained relevance diversity. In this paper, we present a Chinese short video dataset with 4-level relevance annotations, filling a critical resource void. Further, we propose a semi-supervised synthetic data pipeline where two collaboratively trained models generate domain-adaptive short video data with controllable relevance labels. Our method enhances relevance-level diversity by synthesizing samples for underrepresented intermediate relevance labels, resulting in a more balanced and semantically rich training data set.

Extensive offline experiments show that the embedding model trained on our synthesized data outperforms those using data generated based on prompting or vanilla supervised fine-tuning(SFT). Moreover, we demonstrate that incorporating more diverse fine-grained relevance levels in training data enhances the model's sensitivity to subtle semantic distinctions, highlighting the value of fine-grained relevance supervision in embedding learning.  In the search enhanced recommendation pipeline of Douyin's dual-column scenario, through online A/B testing, the proposed model increased click-through rate(CTR) by 1.45\%, raised the proportion of Strong Relevance Ratio (SRR) by 4.9\%, and improved the Image User Penetration Rate (IUPR) by 0.1054\%.

\end{abstract}


\section{Introduction}

Embedding models are fundamental components in many natural language processing applications, including web search, recommendation systems and so on\cite{huang2020embedding,zhao2023embedding,xi2025aug2search}. Recent studies have explored using LLMs to improve embedding quality \cite{nie2024text}, particularly by synthesizing diverse, high-quality training data to enhance the semantic accuracy of embeddings. Gemini Embedding \cite{lee2025gemini} propose a multi-stage prompting strategy to synthesize both retrieval and classification datasets to improve data quality; Qwen3 embedding \cite{zhang2025qwen3} and KaLM-Embedding \cite{hu2025kalm} leverage Persona Hub \cite{ge2024scaling} to guide the data synthesis process, enabling the LLM to generate data from a variety of user personas to construct diverse synthetic train set.

Despite the effectiveness of these methods, existing LLM-based data synthesis methods are predominantly prompt-based and inherently limited by the generative capabilities of the LLM itself. As a result, there remains a noticeable gap between synthetic data and human-curated ground truth data—particularly in low-source domain-specific scenarios such as short video platforms \cite{zhang2024value}. \citet{tang2024we} proposed FinMTEB, a counterpart to MTEB~\cite{muennighoff2022mteb,enevoldsen2502mmteb} that consists of financial domain-specific text datasets, and observed that the state-of-the-art embedding models on MTEB got a significant performance drop on FinMTEB compared to their performance on MTEB. This highlights the necessity of synthesizing data that aligns with domain-specific distributions.

Moreover, current synthesis methods primarily focus on diversity in terms of attributes such as input length, difficulty, and user personas. Little attention has been paid to diversity in terms of semantic relevance, which plays a crucial role in downstream tasks like retrieval and ranking. Most synthesis methods adopt a binary relevance scheme (i.e., relevant vs. irrelevant) and rely on contrastive learning to train embedding models. Although hard negative mining\cite{zhang2022unsupervised} has been widely employed to enhance the model's sensitivity to subtle relevance differences, the binary formulation introduces a misalignment between the training objective and downstream task requirements. For example, retrieval tasks often demand fine-grained relevance ranking of documents rather than coarse binary relevance classification.

To address these limitations, we introduce a short video relevance dataset, along with two test sets designed for evaluating retrieval and pair classification tasks. All of these datasets are annotated with 4-level fine-grained relevance labels. 
Furthermore, we propose a \textbf{Semi-Supervised Relevance-Aware} data synthesis pipeline, termed \textbf{SSRA}. 
SSRA captures domain-specific data distributions while enabling the controllable generation of samples with predefined semantic relevance levels. It comprises two stages: the first enhances diversity across the synthesized data, and the second improves the precision of relevance-conditioned generation. 
By jointly training on annotated domain-specific relevance datasets and high-quality synthetic data generated by SSRA, the embedding model is able to capture fine-grained semantic relevance tailored to the target domain.

We empirically validate SSRA by augmenting training data for the Qwen3-Embedding 4B model. Compared to the baseline trained on annotated data only, our method achieves a 1.73\% improvement in nDCG@10 on the retrieval task and a 2.80\% gain in average precision (AP) on pair classification. SSRA outperforms prompt-based synthetic data generation and vanilla supervised fine-tuning (SFT) synthetic data generation baselines, demonstrating its effectiveness in synthesizing domain-relevant data with target fine-grained relevance label. Moreover, our analysis highlights the importance of fine-grained semantic relevance diversity in improving downstream performance of embedding models. The embedding model trained with SSRA-augmented data has been deployed in the search enhanced recommendation pipeline of a Douyin’s dual-column scenario, where it achieved significant gains in online A/B testing: CTR +1.45\%, SRR +4.9\%, and IUPR +0.1054\%.

In summary, our contributions are as follows:
\begin{itemize}
    \item We propose a short video domain relevance dataset and two corresponding test sets for retrieval and pair classification, filling a critical gap in existing benchmarks.
    \item We introduce \textbf{SSRA}, a novel and effective pipeline for synthesizing domain-specific, fine-grained relevance data for embedding model training.
    \item We empirically demonstrate that increasing the fine-grained semantic relevance diversity of training data improves embedding models’ sensitivity to fine-grained relevance, offering a new perspective on leveraging synthetic data for downstream enhancement.
    \item Extensive experiments show that embedding models trained on SSRA-generated data achieve substantial improvements in both offline metrics on proposed test sets and real-world online deployment, validating the practical effectiveness of our approach.
\end{itemize}

\section{Related Work}
\subsection {Embedding Model Training} 
Embedding models serve as fundamental building blocks for a wide range of natural language processing (NLP) tasks, such as semantic similarity, information retrieval, and classification\cite{chandrasekaran2021evolution,huang2020embedding,keraghel2024beyond}. With the advent of pre-trained language models(PLMs), the learning paradigm of embedding models gradually unify to contrastive learning\cite{wu2020clear,gao2021simcse,izacard2021unsupervised}. Contrastive learning focuses on constructing positive and negative sample pairs, and optimizes the embedding space by pulling positive pairs closer and pushing negative pairs farther apart. A typical loss function used in this framework is the InfoNCE loss\cite{oord2018representation}. 
Recent state-of-the-art LLM-based embedding models\cite{lee2025gemini,zhang2025qwen3}, whose backbone are LLMs\cite{team2023gemini,team2024gemini,yang2025qwen3}, continue to follow the contrastive learning paradigm, focusing on enhancing model performance through higher-quality positive and negative pairs, as well as improved loss functions based on InfoNCE loss.


\subsection {Synthetic Data Generation for Embedding Models} 
Synthetic query generation \cite{liang2020embedding,jeronymo2023inpars} based on given documents has been widely used to create diverse training data for embedding models. \citet{lee2024gecko} showed that documents used to synthesize queries may be suboptimal as positive examples. To address this, their method retrieves candidate documents using the generated query and selects the most relevant one via LLM-based scoring. Moreover, many studies directly synthesize positive and hard negative documents as training data for embedding-based retriever tuning. \citet{kim2025syntriever} leveraged LLMs to synthesize both positive and hard negative documents, aiming to distill the natural language understanding capabilities of LLMs into a lightweight retriever. \citet{shao2025reasonir} proposed a data synthesis pipeline that generates a challenging yet relevant query and a corresponding hard negative document given a document, facilitating the construction of high-quality training triplets. 

While these work improved embedding models by synthesizing high-quality and diverse training data, the notion of relevance is still limited to binary labels. This binary supervision introduces a fundamental discrepancy between the training objective of embedding models and the requirements of downstream tasks such as retrieval and reranking. To overcome this limitation, \citet{esfandiarpoor2025beyond} proposed SyCL, a prompt-based approach to generate 4-level relevance documents for a given query using LLMs. 


\begin{table*}[t]
\centering
\small
\begin{tabular}{l|r|cccc|r|r|c|c}
\toprule
\textbf{Dataset} & \textbf{Size} & \textbf{0} & \textbf{1} & \textbf{2} & \textbf{3} & \textbf{\#Queries} & \textbf{\#Docs} & \textbf{Avg Query Len.} & \textbf{Avg Doc Len.} \\
\midrule
Search Relevance Train Set & 207,439 & 65.47 & 2.85 & 1.60 & 30.07 & 141,950 & 199,677 & 7.47 & 364.69 \\
Retrieval Test Set          & 10,866  & 44.17 & 16.67 & 14.83 & 24.33 & 372     & 10,605  & 9.26 & 454.47 \\
Pair Classification Test Set & 3,390  & 54.45 & 7.17 & 1.95  & 36.43 & 3,291   & 3,389   & 8.14 & 343.93 \\
\bottomrule
\end{tabular}
\caption{Statistics and relevance label distributions of the released datasets. Columns named 0–3 indicate the proportion of each relevance label.}
\label{tab:dataset-statistics}
\end{table*}

\section{Short Video Relevance Dataset}
\label{sec:dataset}
To facilitate the study of domain-specific synthetic data generation with fine-grained relevance control, we present a novel short video relevance dataset tailored to real-world search and recommendation scenarios with 4-level relevance label. Unlike general-purpose datasets, which often lack domain specificity, our dataset captures the rich diversity and complexity of user intent within the short video domain. 

Specifically, we construct a short video search relevance training set along with two evaluation sets, targeting retrieval and pairwise classification tasks respectively. These datasets enables more fine-grained representation learning and evaluation beyond conventional binary relevance setups, and facilitates deeper exploration into semantic matching under domain-specific relevance granularity.

\begin{table}[t]
\centering
\small
\begin{tabular}{c|p{6.5cm}}
\toprule
\textbf{Label} & \textbf{Definition} \\
\midrule
3 & Fully and precisely satisfies the user intent with complete semantic relevance to the query. \\
\midrule
2 & Generally satisfies the user intent and is semantically relevant, but some non-key elements of the content are missing or unmatched. \\
\midrule
1 & Partially satisfies the user intent; relevance is preserved for key entities or concepts, but significant aspects of the need are unmatched. \\
\midrule
0 & Neither satisfies the user intent nor maintains any meaningful semantic relevance to the query. \\
\bottomrule
\end{tabular}
\caption{Definitions of the 4-level relevance labels used for short video search relevance annotation.}
\label{tab:relevance-scale}
\end{table}

\subsection{Data Collection}
\label{data:data_collection}
The core of constructing a search relevance dataset lies in building high-quality query-document pairs. However, in the short video domain, each item is inherently multi-modal and lacks explicit textual representation, making it infeasible to directly collect query-document pairs. To address this, we adopt a two-stage data construction pipeline.

\textbf{Step 1: Query-Item Collection.} We first collect query-item pairs, where each \textit{item} refers to a short video on the short video platform. To obtain high-quality query-item candidates for annotation, we leverage search click logs of Douyin. Specifically, we employ two complementary strategies:
\begin{itemize}
    \item \textbf{Query-driven retrieval:} A set of real user queries is sampled, and candidate items are retrieved using an internal search engine to form query-item pairs.
    \item \textbf{Click-based sampling:} Query-item pairs are directly extracted from search click logs, representing actual user interaction signals.
\end{itemize}

The training set is constructed by combining data from both strategies to ensure richness and diversity. The retrieval test set exclusively uses query-driven retrieval data to simulate realistic candidate ranking scenarios, while the pairwise classification test set is built solely from click-based query-item pairs.

\textbf{Step 2: Document Generation.} Since short videos lack natural language descriptions, we convert each item into a textual representation using a LLM. Specifically, we utilize Doubao-1.5-Pro-32K\footnote{\url{https://console.volcengine.com}} to rewrite the item's available metadata—namely, its OCR\footnote{Optical Character Recognition (OCR) extracts visual text elements, such as overlaid captions or scene text, from video frames to support semantic interpretation.} and ASR\footnote{Automatic Speech Recognition (ASR) transcribes spoken audio in the video into text, enabling downstream language understanding tasks.} fields, into coherent textual descriptions. These rewritten descriptions, combined with the item's title, form the final \textit{document} associated with each item. The prompt for rewrite OCR and ASR can be found in Appendix-Prompts.

It is worth noting that item ranking in search list is influenced by factors beyond semantic relevance, such as personalization and item popularity. As our objective is to model pure semantic relevance, we filter out query-item pairs with extremely low click-through rates to reduce bias from non-relevance signals.

\subsection{Data Annotation}
\label{data:data_annotation}
After selecting query-document pairs, each query-document pair is annotated with a relevance label on a 4-level scale $\{0, 1, 2, 3\}$ based on a unified annotation guideline. The definitions of each relevance level are detailed in Table~\ref{tab:relevance-scale}. This multi-level relevance scheme is specifically designed to reflect nuanced semantic relevance in the context of short video search, capturing domain-specific aspects of content matching and user intent understanding.

To ensure annotation quality and inter-annotator consistency, we adopt a \textbf{dual-annotation with adjudication} protocol. Specifically, each pair is independently annotated by two annotators. In cases where their annotations disagree, an expert examiner—well-versed in short video search relevance—is consulted to adjudicate and assign the final label.

\begin{figure*}[!t]
	\centering
    \includegraphics[scale=0.50]{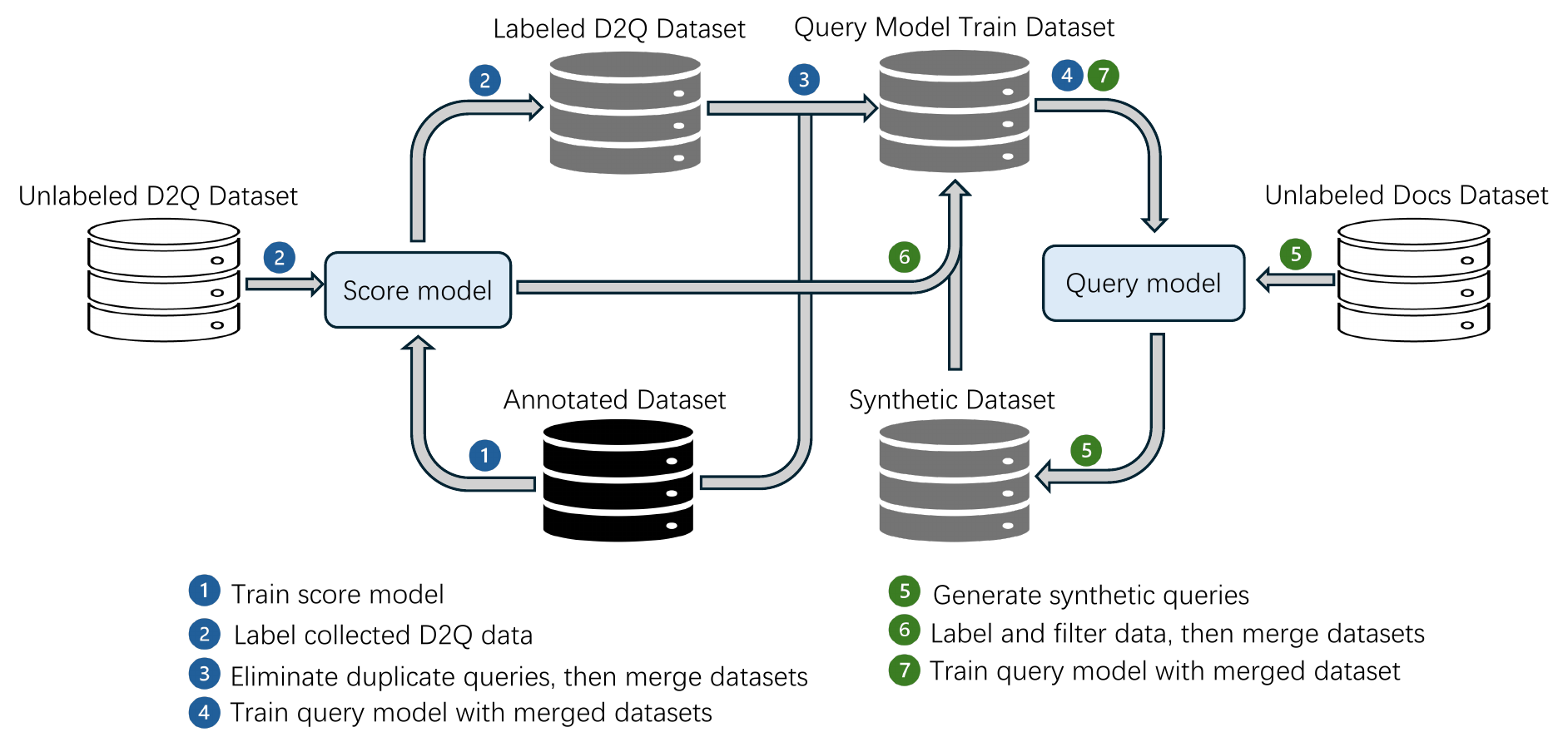}

	\caption{The overview of the proposed two-stage \textbf{Semi-Supervised Relevance-Aware} data synthesis(\textbf{SSRA}) pipeline. Steps 1–4 on the left (highlighted in blue) correspond to Stage 1, while steps 5–7 on the right (highlighted in green) represent Stage 2. The term \textbf{D2Q} refers to a data structure where each document is associated with multiple queries.}
	\label{fig:framework}
\end{figure*}

\subsection{Dataset Statistics}
\label{data:data_statistics}
The overall statistics of the three datasets are presented in Table~\ref{tab:dataset-statistics}. It is worth noting that relevance labels 1 and 2 are significantly underrepresented across all datasets, resulting in a highly imbalanced label distribution along the 4-level relevance scale. This observation highlights the necessity of employing data synthesis techniques to enrich mid-level relevance samples, thereby improving the training signal quality and enabling more effective learning of fine-grained relevance distinctions.

\section{Preliminaries}
\subsection{Problem Formulation}

Let $\mathcal{D}_\text{labeled} = \{(q_i, d_i, s_i)\}_{i=1}^{N}$ be a labeled dataset, where $q_i$ denotes a query, $d_i$ denotes a document, and $s_i \in \mathcal{S} = \{0, 1, 2, 3\}$ represents the discrete relevance label between the query and document. Additionally, we are given an unlabeled dataset $\mathcal{D}_\text{unlabeled} = \{(q_j, d_j)\}_{j=1}^{M}$ containing a large number of query-document pairs without associated relevance labels. The queries in both datasets are assumed to follow a domain-specific query distribution $P_{\text{domain}}$, i.e., $q_i \sim P_{\text{domain}} \forall q_i \in \mathcal{D}_{\text{label}}, \cup \mathcal{D}_{\text{unlabeled}} $.

Our goal is to learn a controllable \textit{query model} $f: (d, s) \mapsto \hat{q}$ that, given a document $d$ and a target relevance label $s \in \mathcal{S}$, generates a query $\hat{q}$ such that:  

1) $\hat{q}$ is sampled from the domain distribution $P_{\text{domain}_i}$, and  

2) the semantic relevance between $\hat{q}$ and $d$ aligns with the specified score $s$.

By leveraging this model, we can synthesize query-document pairs with fine-grained, controllable relevance levels that match the target domain distribution.

\section{Method}
In this section, we present our proposed SSRA pipeline. The overall framework is illustrated in Figure \ref{fig:framework}. Our approach is designed as a two-stage process, each targeting distinct objectives to progressively improve query synthesis quality.

\begin{figure}[!t]
	\centering
    \includegraphics[scale=0.335]{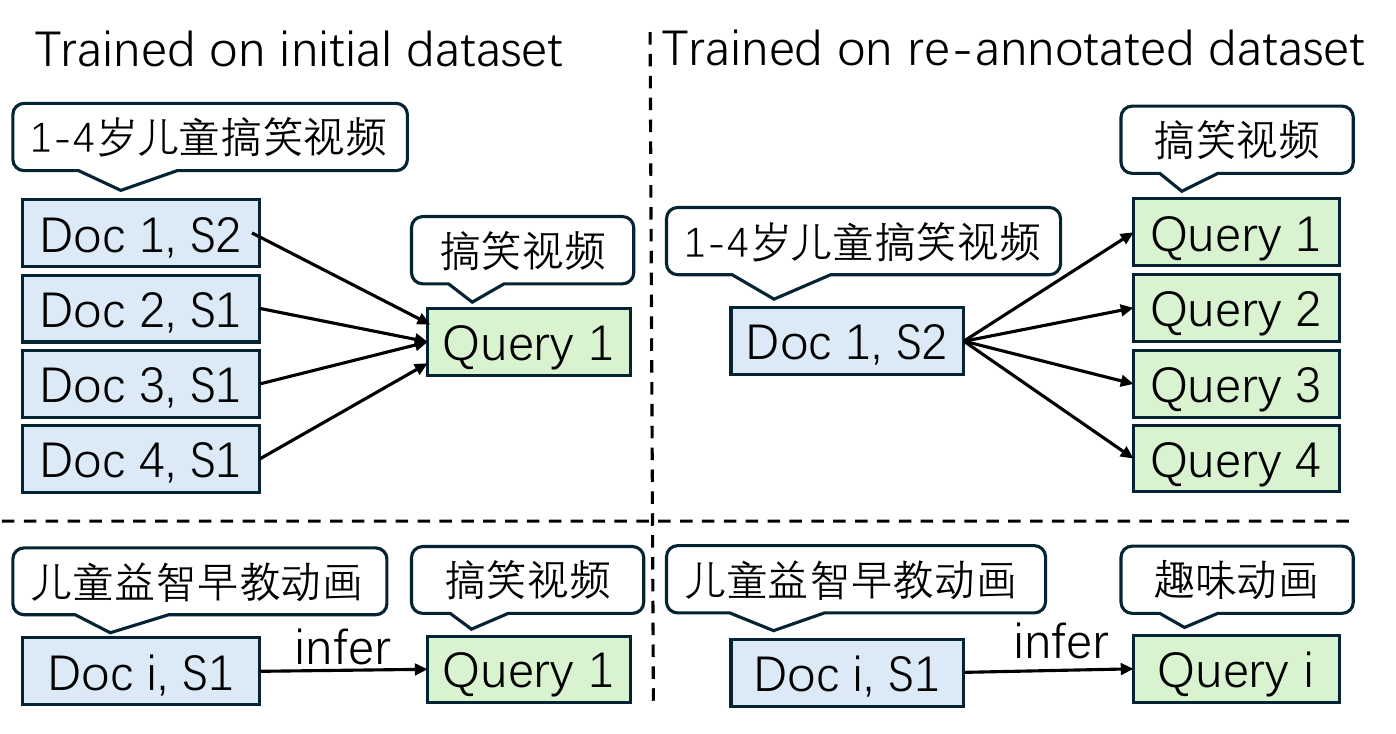}

	\caption{
Illustration of our score-based re-annotation strategy in Stage 1. \textbf{$Si$} represents the target relevance score. 
\textbf{Left:} In the original labeled dataset, multiple documents are associated with the same high-frequency query, leading to reduced diversity in query generation. 
\textbf{Right:} After Stage 1 re-annotation, each document is linked to multiple queries with varying relevance labels, as labeled by a tuned score model. This enables the query generation model to synthesize diverse queries for unseen documents during inference.
}
	\label{fig:stage1}
\end{figure}

\begin{table*}[t]
  \centering
  \small
    \begin{tabular}{l|c|cccc|c|cccc}
    \toprule
    \multirow{2}{*}{\textbf{Method}} 
    & \multicolumn{5}{c|}{\textbf{0.6B Model}} 
    & \multicolumn{5}{c}{\textbf{4B Model}} \\
    
    & nDCG@10 & AP@$\geq$1 & AP@$\geq$2 & AP@$\geq$3 & Avg
    & nDCG@10 & AP@$\geq$1 & AP@$\geq$2 & AP@$\geq$3 & Avg \\
    \midrule
    Base Model (BM)   & 71.36 & 72.63 & 69.12 & 66.74 & 69.50 
                      & 73.20 & 72.96 & 69.14 & 66.61 & 69.57 \\
    \midrule
    SyCL Modified        & 71.50 & 70.34 & 66.90 & 64.75 & 67.33 
                      & 73.56 & 71.30 & 67.43 & 65.11 & 67.95 \\
    Vanilla SFT          & 71.88 & 73.72 & 70.08 & 67.38 & 70.39 
                      & 74.32 & 74.22 & 70.49 & 67.90 & 70.87 \\
    \midrule
    SSRA              & 71.97 & 74.13 & 70.43 & 67.80 & 70.79 
                      & 74.47 & 74.92 & 71.18 & 68.47 & 71.52 \\
    \bottomrule
    \end{tabular}
  \caption{
  Retrieval and pair classification evaluation results for Qwen3-Embedding of different scales (0.6B vs. 4B). AP@$\geq i$ represents Average Precision(AP) computed by treating all samples with relevance labels greater than or equal to $i$ as positive (relevant), and the rest as negative (irrelevant). SSRA consistently improves both nDCG@10 and AP across model sizes.
  }
  \label{tab:main_eval_results}
\end{table*}

\subsection{Stage 1: Enhancing Query Diversity via Score-Based Re-annotation}
\label{SSRA:stage1}
As discussed in a previous section, one source of annotated data in our proposed dataset involves retrieving multiple candidate documents for a given query, followed by manual annotation. This data construction paradigm reduces the diversity of generated queries from \textit{query model}, as the \textit{query model} may learn to map multiple distinct documents to the same query during training.

To address this issue, we introduce a relevance classification model, referred to as the \textit{score model}, trained on the labeled dataset to predict discrete 4 relevance labels. Using this model, we collect unlabeled data grouped by document and assign relevance labels to associated queries, enabling the construction of training instances where each document is paired with multiple queries exhibiting different relevance levels. Finally, we merge the resulting synthetic dataset with the deduplicated version of the original labeled data to train the initial version of \textit{query model}. This restructured training corpus helps enhance query diversity in the early stage of model learning. Moreover, leveraging real user queries extracted from large-scale unlabeled data enables the model to better approximate the distribution of in-domain queries, thus fostering more realistic and domain-specific query synthesis. Figure \ref{fig:stage1} shows the change during inference after applying the re-labeled training dataset for \textit{query model}.

\subsection{Stage 2: Enhancing Alignment Between Synthesized Queries and Target Relevance Labels}
\label{SSRA:stage2}
Despite being trained on carefully constructed data, the query generation model still struggles to reliably synthesize queries that align with specified target relevance labels. 

To address this issue while maintaining training efficiency, we adopt an iterative refinement approach that augments the training data with generated queries that better match the desired relevance labels. This is achieved through the following steps:

\begin{itemize}
    \item \textbf{Initial Synthesis:} We first leverage the \textit{query model} trained in Stage 1 to generate a large pool of synthetic queries on the unlabeled document set, conditioned on specific relevance labels.
    
    \item \textbf{Score-based Filtering:} To ensure the quality of synthesized data, we apply the \textit{score model} tuned on proposed labeled data set to predict the relevance label of each query-document pair. Only those synthetic examples where the predicted label matches the intended target label are retained.

    \item \textbf{Pairwise Consistency Filtering via LLM:} Given that the score model may still introduce errors, we further refine the dataset using a LLM. For each document, we compare pairs of generated queries with differing relevance labels, and filter out cases where the relative order of relevance—judged by the LLM—is inconsistent with the relevance labels. This pairwise filtering step improves the fidelity of relevance differentiation within a single document context.
\end{itemize}

The remaining high-quality samples are then combined with the original training data from Stage 1 to form an enriched dataset for a second round of query model training. This iterative training process leads to a more robust and relevance-aligned query generator.

\begin{table*}[t]
  \centering
    \begin{tabular}{l|cccc|cccc}
    \toprule
    \multirow{2}[2]{*}{\textbf{{Method}}} & \multicolumn{4}{c|}{\textbf{0.6B Model}} & \multicolumn{4}{c}{\textbf{4B Model}}  \\
     & nDCG@10 & AP@$\geq$ 1 & AP@$\geq$ 2 & AP@$\geq$ 3 &nDCG@10 & AP@$\geq$ 1 & AP@$\geq$ 2 & AP@$\geq$ 3 \\ 
    \midrule
    Binary & 71.66 & 87.36 & 75.54 & 54.22 & 73.73 & 91.19 & 77.02 & 58.39 \\
    Multi-Level & 71.78 & 88.89 & 76.70 & 52.86  & 74.23 & 91.28 & 77.63 & 55.25  \\
    \bottomrule
    \end{tabular}
      \caption{Performance of Qwen3-embedding trained with SSRA augmented data with binary({0,1}) and multi-level({0,1,2,3}) relevance labels. Note that the AP results are calculated in a balanced-version pair classification test set.}
  \label{tab:binary_multi}%
\end{table*}%

\section{Experiments}
\subsection{Experimental Settings}

\subsubsection{Implementation Details.} To train both the \textit{query model} and the \textit{score model}, we adopt Doubao-1.5-Pro-32K as the backbone and both models are trained using a batch size of 16 and an initial learning rate of $1 \times 10^{-5}$. To utilize 4-level relevance labels, we utilize infoNCE loss with label as additional weight of loss\footnote{$\mathcal{L}=\frac{1}{B}\sum_{i=1}^{B}s_i \cdot \mathcal{L}_{infoNCE}$, where $s_i \in \{0,1,2,3\}$ is the $i$-th relevance label in batch, B is batch size.}.

Following the two-stage SSRA pipeline, we use the final \textit{query model} to synthesize relevance-controlled queries for 1 million distinct short video documents, covering relevance labels of 1, 2, and 3\footnote{We use in-batch negative method to get negative pairs, whose relevance label is 0.}. The synthesized queries are then filtered using the score model to ensure label consistency. These high-quality synthetic queries are combined with the proposed annotated train set to form the final train set for the embedding model. 

We employ Qwen3-Embedding\footnote{\url{https://huggingface.co/Qwen}} as the backbone for embedding model training, and utilize both in-batch negative sampling and hard negative mining techniques. For  We train embedding models at two parameter scales: 0.6B and 4B. For both settings, we adopt the Qwen3-Embedding architecture with lora fine-tuning, following the implementation from the Qwen3-Embedding blog\footnote{\url{https://qwenlm.github.io/blog/qwen3-embedding/}}. Specifically, we apply LoRA~\cite{hu2022lora} with a rank of 32 and the models are trained with a batch size of 512 and a learning rate of $1 \times 10^{-5}$.

More implementation details and corresponding codes are provided in Appendix-Implementation.

\subsubsection{Evaluation Settings.} 
We evaluate our methods on the proposed two benchmarks in short video domain: a retrieval test set and a pairwise classification test set. Each benchmark is annotated with four-level relevance labels (0 to 3), capturing fine-grained relevance signals. More details of proposed three data sets can be found in Appendix-Dataset.

For the retrieval test set, we follow the MTEB\cite{muennighoff2022mteb} evaluation protocol and report nDCG@10 as the primary metric. For the pairwise classification test set, we also follow the MTEB standard and report Average Precision (AP) at different relevance thresholds (label $\geq$ 1, 2, and 3) to assess semantic matching quality at varying levels of granularity.

\subsubsection{Baselines.} 
We consider several baselines to evaluate the effectiveness of our SSRA pipeline. First, we include a \textbf{Base Model} trained directly on the annotated training data without any synthetic augmentation. Second, we evaluate a \textbf{Vanilla SFT} baseline, where synthetic data is generated by a query model trained on the annotated dataset and used to augment training dataset. Third, we include a prompt-based baseline, \textbf{SyCL Modified}, adapted from SyCL\cite{esfandiarpoor2025beyond} which originally generates documents of varying relevance labels given a query. We retain its core prompt design but modify the objective to generate relevance-controlled queries given a document, making it a representative of prompt-based query synthesis methods.

\subsection{Main Results}
Table \ref{tab:main_eval_results} shows our main results on the effectiveness of SSRA-generated, domain-specific synthetic data with controllable fine-grained relevance levels for training embedding models. 
Compared to the Base Model, our method yields a 1.73\% improvement in nDCG@10 on the retrieval task and a 2.80\% gain in average precision (AP) on the pair classification task for the Qwen3-Embedding 4B model. For the smaller Qwen3-Embedding 0.6B model, our approach still brings notable gains, with a 0.85\% improvement in nDCG@10 and a 1.86\% increase in AP.

Our method consistently enhances the performance of Base Model on domain-specific retrieval and pairwise classification tasks. The improvements are observed across models of different parameter scales, demonstrating the robustness of our approach. 
Furthermore, it outperforms both prompt-based generation methods and direct supervised fine-tuning (SFT) baselines, highlighting the effectiveness of our relevance-controllable query synthesis framework.

Notably, the prompt-based synthesis approach \textbf{SyCL Modified} not only fails to improve, but in fact degrades the performance of the embedding model on the pairwise classification task. Specifically, it leads to a drop of 2.17 points in average AP for the 0.6B model and 1.62 points for the 4B model. In contrast, the tuning-based synthesis methods, even when simply trained on annotated data, contributes positively to model performance. Surprisingly, our SSRA method yields substantial gains, improving average AP by 1.29 points on the 0.6B model and 1.95 points on the 4B model. These results highlight the limitations of prompt-based approaches in generating effective domain-specific training data.

\subsection{Ablation Study and Additional Analysis}

We conducted additional experiments, aiming at answering the following key research questions (RQs).
\subsubsection{RQ1: Does fine-grained relevance with balanced distribution matters?}

To assess whether a more diverse relevance distribution (i.e., 4-level relevance labels with a more balanced label distribution) can lead to improved model performance compared to traditional binary supervision, we conduct controlled experiments using the SSRA-augmented dataset. We evaluate our model using the proposed retrieval test set and a subset of the pair classification task. Since the original pair classification subset contains an imbalanced distribution with only a small proportion of label 1 and 2 samples, we construct a balanced version consisting of 60 test instances for each label (0, 1, 2, 3).

As shown in Table \ref{tab:dataset-statistics}, in the original human-annotated dataset, samples with the middle relevance label (12) are underrepresented, and conventional binary annotation schemes tend to label partially matched pairs as non-relevant (i.e., score 0), resulting in a skewed and coarse relevance distribution. In contrast, the SSRA framework, via the \textit{query model}, synthesizes additional middle-relevance samples, leading to a more balanced and fine-grained label distribution. 

To simulate the binary relevance setting, we construct a binary subset by filtering out all score-12 samples from the SSRA-augmented dataset. The multi-level subset retains the full range of 4-level relevance labels. We train identical models on both subsets for 2,000 steps under the same training configurations. The final evaluation results are reported in Table \ref{tab:binary_multi}.

We find that, for the retrieval task, training data with intermediate relevance levels leads to a notable improvement in nDCG@10, highlighting the importance of fine-grained and diverse relevance labels in enhancing retrieval performance. 

For the pair classification task, incorporating more training data with intermediate relevance levels can improve the model's ability to distinguish the middle relevance cases with an improvement on  AP@$\geq$1 and AP@$\geq$2. However, this may come at the cost of reduced performance in distinguishing the most relevant pairs from those of lower relevance levels.
This finding suggests that while fine-grained relevance annotations are not always equally beneficial across all pair classification thresholds, they remain essential for tasks that require nuanced differentiation, particularly those involving low-relevance or borderline cases.

\begin{table}[!t]
  \centering
    \begin{tabular}{l|cc|cc}
    \toprule
    \multirow{2}[2]{*}{\textbf{Method}} & \multicolumn{2}{c|}{\textbf{0.6B Model}} & \multicolumn{2}{c}{\textbf{4B Model}}  \\
     & nDCG@10 & AP &nDCG@10 & AP \\ 
    \midrule
    w/o s1\&s2 & 71.44 & 70.70 & 74.02 & 70.34 \\
    w/o s2 & 71.79 & 70.77 & 74.23 & 71.30 \\
    SSRA & 71.97 & 74.13 & 74.47 & 74.92 \\
    \bottomrule
    \end{tabular}
      \caption{Ablation results of the two-stage SSRA pipeline using Qwen3-embedding. 
      }
  \label{tab:ablation}%
\end{table}%

\subsubsection{RQ2: How Do the Two Stages of SSRA Contribute to Retrieval and Pairwise Classification Tasks?}

To investigate the contributions of the two-stage SSRA framework across different tasks, we conduct a series of ablation studies, as reported in Table \ref{tab:ablation}. The results reveal that both stages contribute to performance gains in the retrieval task, whereas the improvement in pairwise classification primarily stems from the second stage.

To further analyze the underlying causes of this difference, we examine the impact of each stage on the \textit{query model} from two distinct perspectives:

\paragraph{Effect of Stage~1 on Query Diversity.} To quantify the effect of Stage 1 on query diversity, we compare the proportion of duplicate queries generated by two models: (i) a baseline model trained directly on annotated data, and (ii) a model trained with re-labeled data from Stage 1 of SSRA. Specifically, both models generate queries of relevance levels 1, 2, and 3 for 10{,}000 sampled documents. We then compute the duplicate rate by identifying exact matches among the generated queries.

We observe that the baseline model produces duplicate queries at a rate of 6.57\%, whereas the Stage 1-enhanced model reduces this rate to 5.20\%, corresponding to a 20.85\% relative decrease. This result demonstrates that SSRA Stage~1 effectively improves query diversity. Detailed statistics of query frequencies can be found in the Appendix-Additional Experimental Results.

\paragraph{Effect of Stage~2 on Relevance Alignment.} To evaluate whether Stage 2 training enhances the controllability of the query generation model—specifically its ability to synthesize queries that match a target relevance label—we design the following experiment. 

We randomly sample 200 documents and use query generation models trained with only Stage 1 and with both Stage 1 and Stage 2 to generate queries intended to match relevance levels 1, 2, and 3, respectively. The generated query-document pairs are then annotated by human annotators using the same 4-level relevance schema. We record the number of generated queries whose annotated relevance label matches the intended label.


As shown in Table~\ref{tab:query-generation-eval}, the model trained with Stage 2 significantly improves the alignment between intended and actual relevance levels with a 25.43\% relative improvement in consistency, demonstrating enhanced controllability and generation precision.

\begin{table}[t]
\centering
\begin{tabular}{l|cc}
\toprule
\textbf{Relevance Label} & \textbf{Stage-1 Only} & \textbf{Stage-1 + Stage-2} \\
\midrule
1 & 81 / 200 & 130 / 200 \\
2 & 80 / 200 & 131 / 200 \\
3 & 189 / 200 & 178 / 200 \\
\bottomrule
\end{tabular}
\caption{
Number of synthesized queries matching the intended relevance labels across training stages.
}
\label{tab:query-generation-eval}
\end{table}

\paragraph{Interpretation of Task-Specific Gains.} Based on the above analyses, we interpret the task-specific contributions of SSRA as follows:

For the retrieval task, the model is required to select a subset of relevant documents from a large candidate pool given a query. As such, it is sensitive not only to fine-grained relevance, but also to the diversity of training data. Improved diversity can help mitigate long-tail distribution issues and improve generalization. Therefore, the retrieval task benefits from both stages of SSRA.

In contrast, the pairwise classification task focuses solely on determining the relevance of individual query-document pairs, and does not heavily rely on query diversity. Consequently, it mainly benefits from the second stage, which emphasizes fine-grained relevance supervision.

\begin{table}[h]
\centering
\begin{tabular}{lccc}
\toprule
\textbf{Metric} & \textbf{CTR} & \textbf{SRR} & \textbf{IUPR} \\
\midrule
\textbf{Relative Gain} & +1.45\% & +4.90\% & +0.1054\% \\
\bottomrule
\end{tabular}
\caption{Relative improvement in online A/B testing metrics after deploying the SSRA-based embedding model.}
\label{tab:abtest}
\end{table}

\section{Online A/B Testing}
Based on the embedding model trained with synthetic data from SSRA pipeline, we conducted rigorous online A/B tests in the dual-column scenario of Douyin from May 9 to May 19, 2025, with hundreds of millions of users participating per day. Users were randomly assigned to control and experimental groups via hash bucketing (190 million users each). The embedding model is used to filter out videos irrelevant to the query, ensuring that the displayed videos are closely related to users' historical search queries. 


The results are presented in Table \ref{tab:abtest}, focusing on several key metrics: CTR (Click-Through Rate), SRR (Strong Relevance Ratio), and IUPR (Image User Penetration Rate). SRR is defined as the percentage of content strongly relevant to the user's search query among the videos ultimately recommended. IUPR refers to the proportion of users who clicked on image content among the total daily active users, which is a core metric in the dual-column scenario.

\section{Conclusion}
In this work, we address the limitations of existing LLM-based data synthesis approaches in capturing domain-specific fine-grained semantic relevance. We introduce a relevance dataset in the short video domain and propose \textbf{SSRA}, a semi-supervised synthesis pipeline that enables controllable generation of relevance-annotated data aligned with domain distributions. Empirical results demonstrate that SSRA consistently improves embedding model performance on both retrieval and pair classification tasks, outperforming prompt-based and vanilla supervised fine-tuning baselines. Our findings underscore the importance of incorporating fine-grained semantic diversity in synthetic data and point toward promising directions for enhancing embedding models through relevance-aware data generation.


\section{Acknowledgments}
We sincerely appreciate our colleagues at ByteDance for their support. They contributed to this work but were not listed as authors. All contributors are listed in alphabetical order by last name: Zirui Guo, Jianheng Ma, Yaling Mou,  Lu Ruan, Hailiang Wang.

\bibliography{aaai2026}

@inproceedings{huang2020embedding,
  title={Embedding-based retrieval in facebook search},
  author={Huang, Jui-Ting and Sharma, Ashish and Sun, Shuying and Xia, Li and Zhang, David and Pronin, Philip and Padmanabhan, Janani and Ottaviano, Giuseppe and Yang, Linjun},
  booktitle={Proceedings of the 26th ACM SIGKDD International Conference on Knowledge Discovery \& Data Mining},
  pages={2553--2561},
  year={2020}
}

@article{zhao2023embedding,
  title={Embedding in recommender systems: A survey},
  author={Zhao, Xiangyu and Wang, Maolin and Zhao, Xinjian and Li, Jiansheng and Zhou, Shucheng and Yin, Dawei and Li, Qing and Tang, Jiliang and Guo, Ruocheng},
  journal={arXiv preprint arXiv:2310.18608},
  year={2023}
}

@article{zhang2025qwen3,
  title={Qwen3 Embedding: Advancing Text Embedding and Reranking Through Foundation Models},
  author={Zhang, Yanzhao and Li, Mingxin and Long, Dingkun and Zhang, Xin and Lin, Huan and Yang, Baosong and Xie, Pengjun and Yang, An and Liu, Dayiheng and Lin, Junyang and others},
  journal={arXiv preprint arXiv:2506.05176},
  year={2025}
}

@article{yang2025qwen3,
  title={Qwen3 technical report},
  author={Yang, An and Li, Anfeng and Yang, Baosong and Zhang, Beichen and Hui, Binyuan and Zheng, Bo and Yu, Bowen and Gao, Chang and Huang, Chengen and Lv, Chenxu and others},
  journal={arXiv preprint arXiv:2505.09388},
  year={2025}
}

@article{hu2025kalm,
  title={Kalm-embedding: Superior training data brings a stronger embedding model},
  author={Hu, Xinshuo and Shan, Zifei and Zhao, Xinping and Sun, Zetian and Liu, Zhenyu and Li, Dongfang and Ye, Shaolin and Wei, Xinyuan and Chen, Qian and Hu, Baotian and others},
  journal={arXiv preprint arXiv:2501.01028},
  year={2025}
}

@article{zhang2024value,
  title={The value of AI-generated metadata for UGC platforms: Evidence from a large-scale field experiment},
  author={Zhang, Xinyi and Sun, Chenshuo and Zhang, Renyu and Goh, Khim-Yong},
  journal={arXiv preprint arXiv:2412.18337},
  year={2024}
}

@article{lee2025gemini,
  title={Gemini embedding: Generalizable embeddings from gemini},
  author={Lee, Jinhyuk and Chen, Feiyang and Dua, Sahil and Cer, Daniel and Shanbhogue, Madhuri and Naim, Iftekhar and {\'A}brego, Gustavo Hern{\'a}ndez and Li, Zhe and Chen, Kaifeng and Vera, Henrique Schechter and others},
  journal={arXiv preprint arXiv:2503.07891},
  year={2025}
}

@article{team2023gemini,
  title={Gemini: a family of highly capable multimodal models},
  author={Team, Gemini and Anil, Rohan and Borgeaud, Sebastian and Alayrac, Jean-Baptiste and Yu, Jiahui and Soricut, Radu and Schalkwyk, Johan and Dai, Andrew M and Hauth, Anja and Millican, Katie and others},
  journal={arXiv preprint arXiv:2312.11805},
  year={2023}
}

@article{team2024gemini,
  title={Gemini 1.5: Unlocking multimodal understanding across millions of tokens of context},
  author={Team, Gemini and Georgiev, Petko and Lei, Ving Ian and Burnell, Ryan and Bai, Libin and Gulati, Anmol and Tanzer, Garrett and Vincent, Damien and Pan, Zhufeng and Wang, Shibo and others},
  journal={arXiv preprint arXiv:2403.05530},
  year={2024}
}

@article{ge2024scaling,
  title={Scaling synthetic data creation with 1,000,000,000 personas},
  author={Ge, Tao and Chan, Xin and Wang, Xiaoyang and Yu, Dian and Mi, Haitao and Yu, Dong},
  journal={arXiv preprint arXiv:2406.20094},
  year={2024}
}

@article{liang2020embedding,
  title={Embedding-based zero-shot retrieval through query generation},
  author={Liang, Davis and Xu, Peng and Shakeri, Siamak and Santos, Cicero Nogueira dos and Nallapati, Ramesh and Huang, Zhiheng and Xiang, Bing},
  journal={arXiv preprint arXiv:2009.10270},
  year={2020}
}

@article{jeronymo2023inpars,
  title={Inpars-v2: Large language models as efficient dataset generators for information retrieval},
  author={Jeronymo, Vitor and Bonifacio, Luiz and Abonizio, Hugo and Fadaee, Marzieh and Lotufo, Roberto and Zavrel, Jakub and Nogueira, Rodrigo},
  journal={arXiv preprint arXiv:2301.01820},
  year={2023}
}

@article{lee2024gecko,
  title={Gecko: Versatile text embeddings distilled from large language models},
  author={Lee, Jinhyuk and Dai, Zhuyun and Ren, Xiaoqi and Chen, Blair and Cer, Daniel and Cole, Jeremy R and Hui, Kai and Boratko, Michael and Kapadia, Rajvi and Ding, Wen and others},
  journal={arXiv preprint arXiv:2403.20327},
  year={2024}
}

@article{kim2025syntriever,
  title={Syntriever: How to train your retriever with synthetic data from llms},
  author={Kim, Minsang and Baek, Seungjun},
  journal={arXiv preprint arXiv:2502.03824},
  year={2025}
}

@article{shao2025reasonir,
  title={ReasonIR: Training Retrievers for Reasoning Tasks},
  author={Shao, Rulin and Qiao, Rui and Kishore, Varsha and Muennighoff, Niklas and Lin, Xi Victoria and Rus, Daniela and Low, Bryan Kian Hsiang and Min, Sewon and Yih, Wen-tau and Koh, Pang Wei and others},
  journal={arXiv preprint arXiv:2504.20595},
  year={2025}
}

@article{esfandiarpoor2025beyond,
  title={Beyond Contrastive Learning: Synthetic Data Enables List-wise Training with Multiple Levels of Relevance},
  author={Esfandiarpoor, Reza and Zerveas, George and Zhang, Ruochen and Mgonzo, Macton and Eickhoff, Carsten and Bach, Stephen H},
  journal={arXiv preprint arXiv:2503.23239},
  year={2025}
}

@article{tang2024we,
  title={Do we need domain-specific embedding models? An empirical investigation},
  author={Tang, Yixuan and Yang, Yi},
  journal={arXiv preprint arXiv:2409.18511},
  year={2024}
}

@article{hu2022lora,
  title={Lora: Low-rank adaptation of large language models.},
  author={Hu, Edward J and Shen, Yelong and Wallis, Phillip and Allen-Zhu, Zeyuan and Li, Yuanzhi and Wang, Shean and Wang, Lu and Chen, Weizhu and others},
  journal={ICLR},
  volume={1},
  number={2},
  pages={3},
  year={2022}
}

@article{muennighoff2022mteb,
  title={Mteb: Massive text embedding benchmark},
  author={Muennighoff, Niklas and Tazi, Nouamane and Magne, Lo{\"\i}c and Reimers, Nils},
  journal={arXiv preprint arXiv:2210.07316},
  year={2022}
}

@article{enevoldsen2502mmteb,
  title={Mmteb: Massive multilingual text embedding benchmark (2025)},
  author={Enevoldsen, K and others},
  journal={arXiv preprint arXiv:2502.13595},
  year={2025}
}

@article{chandrasekaran2021evolution,
  title={Evolution of semantic similarity—a survey},
  author={Chandrasekaran, Dhivya and Mago, Vijay},
  journal={Acm Computing Surveys (Csur)},
  volume={54},
  number={2},
  pages={1--37},
  year={2021},
  publisher={ACM New York, NY, USA}
}

@inproceedings{keraghel2024beyond,
  title={Beyond words: a comparative analysis of LLM embeddings for effective clustering},
  author={Keraghel, Imed and Morbieu, Stanislas and Nadif, Mohamed},
  booktitle={International Symposium on Intelligent Data Analysis},
  pages={205--216},
  year={2024},
  organization={Springer}
}

@article{wu2020clear,
  title={Clear: Contrastive learning for sentence representation},
  author={Wu, Zhuofeng and Wang, Sinong and Gu, Jiatao and Khabsa, Madian and Sun, Fei and Ma, Hao},
  journal={arXiv preprint arXiv:2012.15466},
  year={2020}
}

@article{gao2021simcse,
  title={Simcse: Simple contrastive learning of sentence embeddings},
  author={Gao, Tianyu and Yao, Xingcheng and Chen, Danqi},
  journal={arXiv preprint arXiv:2104.08821},
  year={2021}
}

@article{izacard2021unsupervised,
  title={Unsupervised dense information retrieval with contrastive learning},
  author={Izacard, Gautier and Caron, Mathilde and Hosseini, Lucas and Riedel, Sebastian and Bojanowski, Piotr and Joulin, Armand and Grave, Edouard},
  journal={arXiv preprint arXiv:2112.09118},
  year={2021}
}

@article{oord2018representation,
  title={Representation learning with contrastive predictive coding},
  author={Oord, Aaron van den and Li, Yazhe and Vinyals, Oriol},
  journal={arXiv preprint arXiv:1807.03748},
  year={2018}
}

@inproceedings{zhang2022unsupervised,
  title={Unsupervised sentence representation via contrastive learning with mixing negatives},
  author={Zhang, Yanzhao and Zhang, Richong and Mensah, Samuel and Liu, Xudong and Mao, Yongyi},
  booktitle={Proceedings of the AAAI Conference on Artificial Intelligence},
  volume={36},
  number={10},
  pages={11730--11738},
  year={2022}
}

@article{nie2024text,
  title={When text embedding meets large language model: a comprehensive survey},
  author={Nie, Zhijie and Feng, Zhangchi and Li, Mingxin and Zhang, Cunwang and Zhang, Yanzhao and Long, Dingkun and Zhang, Richong},
  journal={arXiv preprint arXiv:2412.09165},
  year={2024}
}

@article{xi2025aug2search,
  title={Aug2Search: Enhancing Facebook Marketplace Search with LLM-Generated Synthetic Data Augmentation},
  author={Xi, Ruijie and Ba, He and Yuan, Hao and Agrawal, Rishu and Tian, Yuxin and Kong, Ruoyan and Prakash, Arul},
  journal={arXiv preprint arXiv:2505.16065},
  year={2025}
}


\appendix
\section{Appendix}
\subsection{Prompts}
This section provides the detailed prompting strategies used throughout our data construction pipeline and the SSRA framework. 

\begin{figure}[!t]
	\centering
    \includegraphics[scale=0.6]{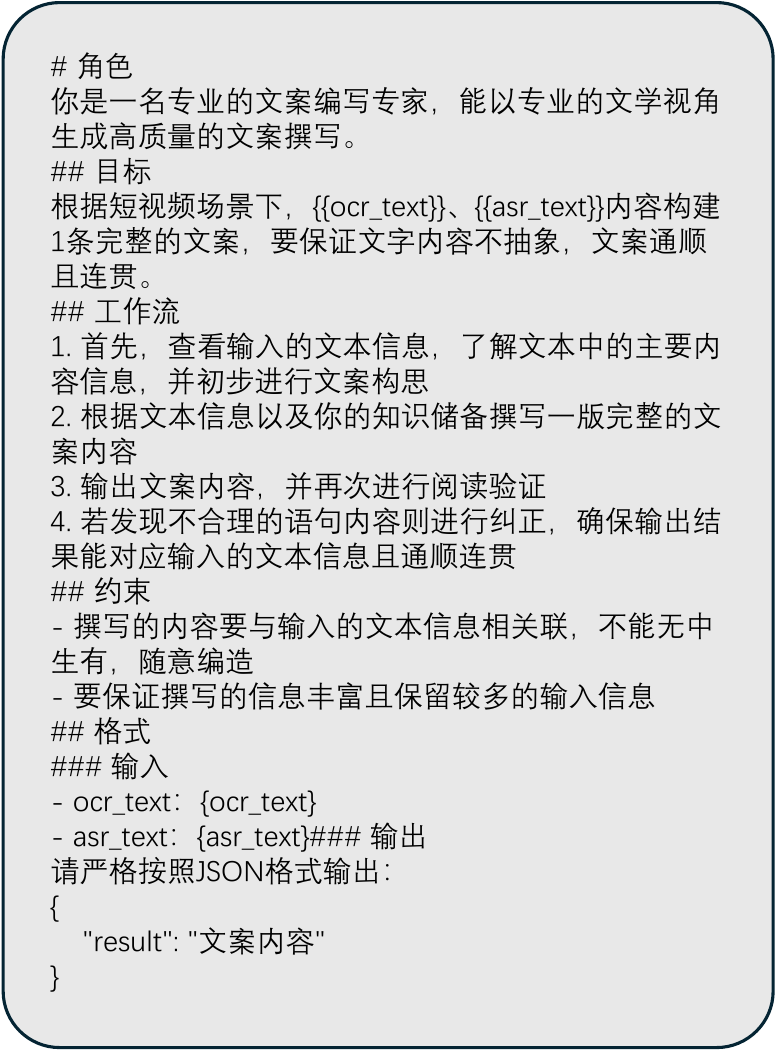}

	\caption{Prompt for short video content rewriting.}
	\label{fig:prompt1}
\end{figure}

\paragraph{Short Video Content Rewriting Prompt.}
To convert multimodal video signals into coherent textual descriptions, we design a prompt that takes OCR and ASR outputs as input and synthesizes a unified document-level representation. This transformation is critical for constructing text-based query-document pairs and the prompt is shown in Figure \ref{fig:prompt1}.

\begin{figure}[!t]
	\centering
    \includegraphics[scale=0.6]{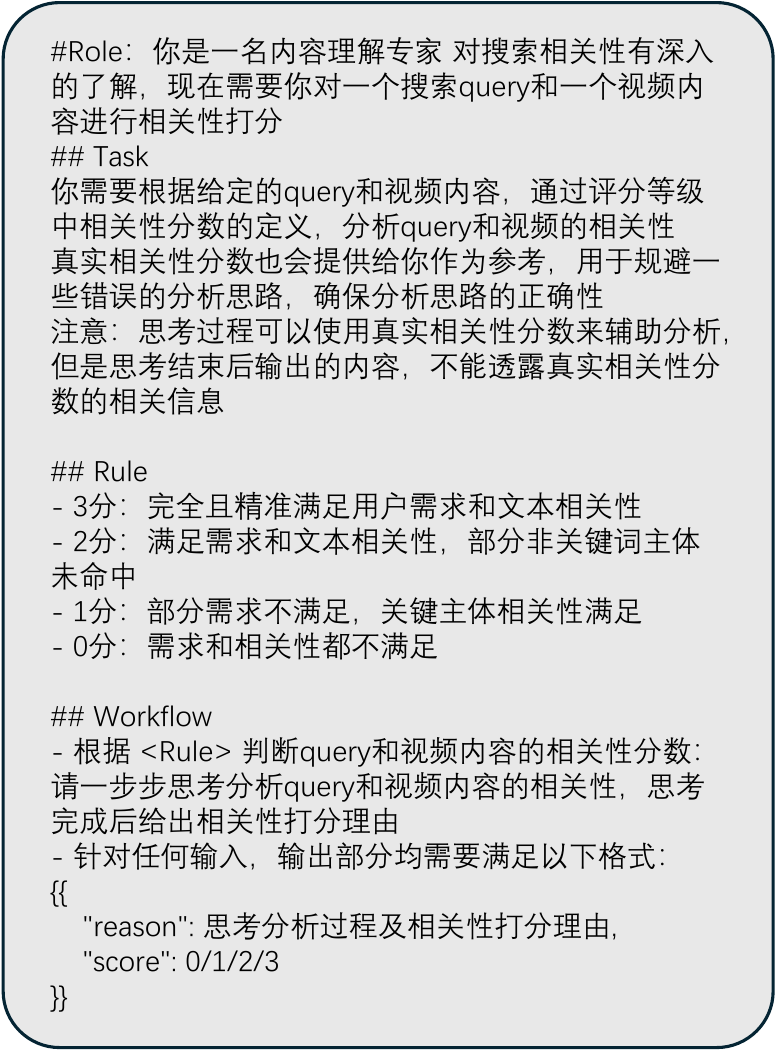}

	\caption{Prompt for score model relevance reasoning data construction.}
	\label{fig:prompt2}
\end{figure}

\paragraph{Two-Stage Prompting for Relevance Reasoning Data Construction.}
To align the score model with human-like relevance reasoning, we adopt a two-stage prompting framework: the model is first guided to conduct step-by-step analysis of the input, followed by a justification and a final relevance score prediction. 

However, our annotated dataset only contains scalar relevance labels without intermediate reasoning steps. To address this, we leverage the DeepSeek-R1\cite{guo2025deepseek} model to generate full reasoning chains. Specifically, given a query-document pair and its associated relevance label, we apply the prompt design illustrated in Figure~\ref{fig:prompt2} to generate the missing analytical reasoning. This enables the construction of a training dataset tailored for two-stage relevance modeling.



\begin{figure}[!t]
	\centering
    \includegraphics[scale=0.6]{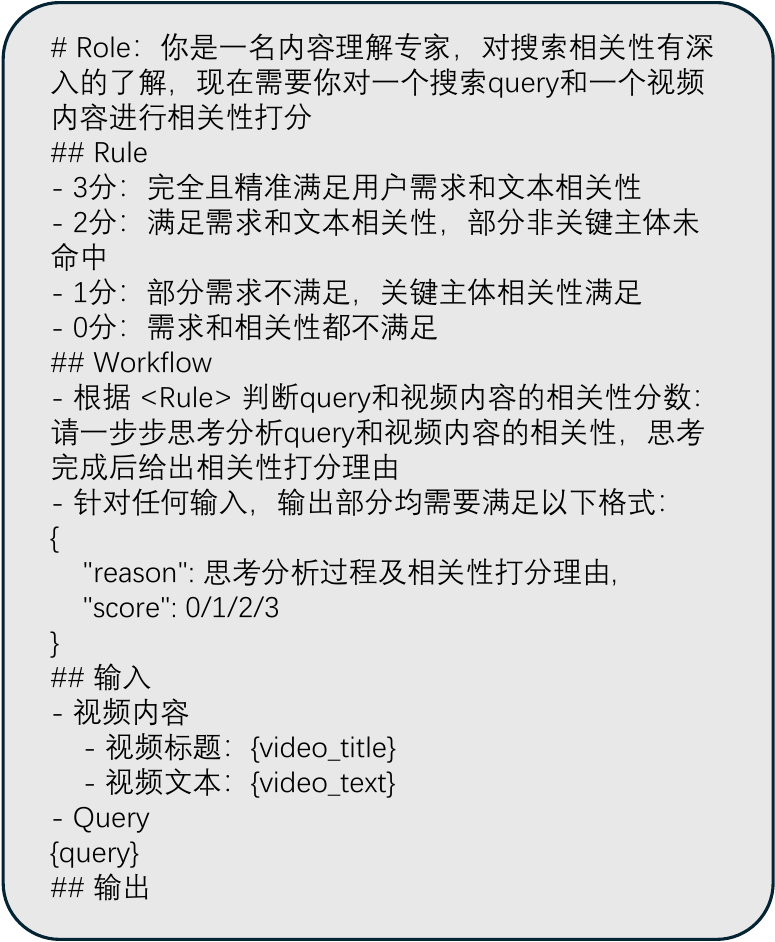}

	\caption{Prompt for score model training and inference.}
	\label{fig:prompt3}
\end{figure}

\paragraph{Score Model Training/Inference Prompt.}
The following prompt in Figure \ref{fig:prompt3} is used for both training and inference of the score model. It guides the model to perform fine-grained relevance assessment between a given query and a video content pair, based on a predefined 4-level relevance schema.

\begin{figure}[!t]
	\centering
    \includegraphics[scale=0.6]{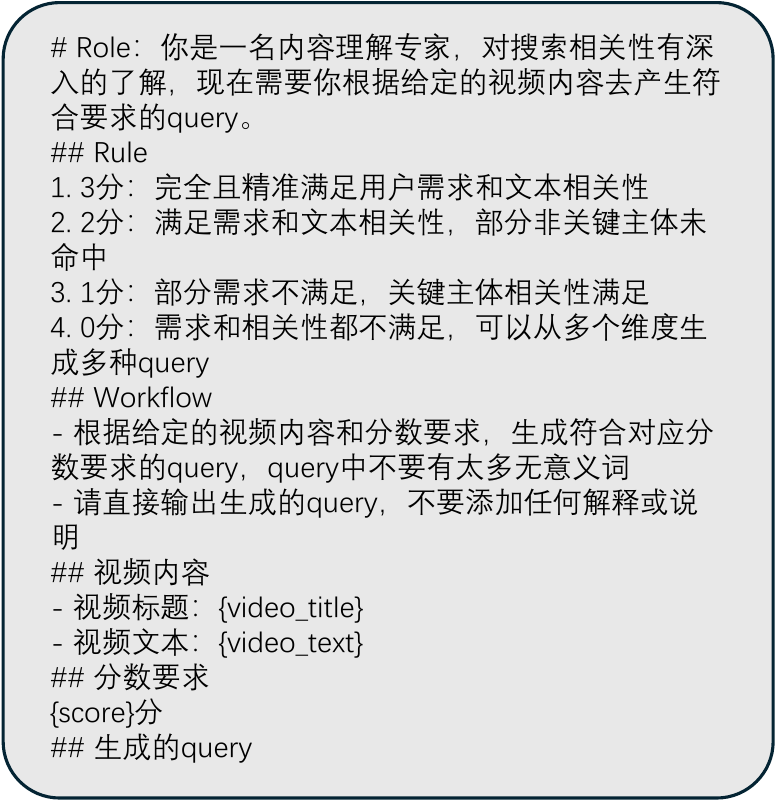}

	\caption{Prompt for query model training and inference.}
	\label{fig:prompt4}
\end{figure}

\paragraph{Query Model Training/Inference Prompt.}
The following prompt in Figure \ref{fig:prompt4} is used to guide the query model to generate queries with a specified relevance level, given the video content. It is utilized in both training (data synthesis) and inference stages.

\begin{figure}[!t]
	\centering
    \includegraphics[scale=0.6]{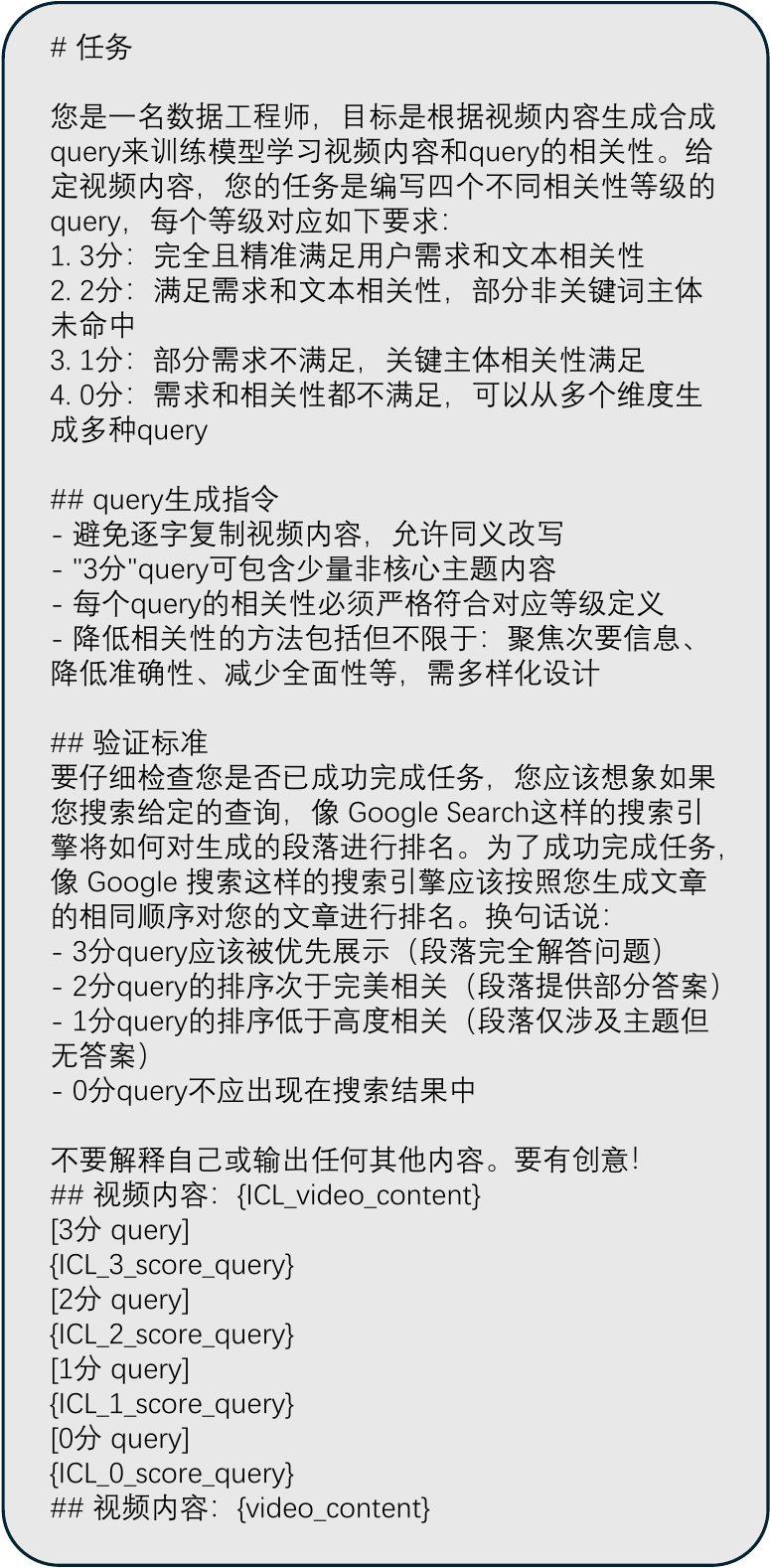}

	\caption{Prompt for SyCL-based query synthesis.}
	\label{fig:prompt5}
\end{figure}

\paragraph{SyCL-based Query Synthesis Prompt.} To synthesize queries with different relevance levels, we design the following prompt shown in Figure \ref{fig:prompt5} following SyCL framework as prompt-based synthesis methods baseline. 

\subsection{Implementation}
\subsubsection{Score model and Query model training}
The key hyper-parameters used for training the score model and the query generation model are summarized in Table~\ref{tab:hyperparams}.

\begin{table}[h]
\centering
\begin{tabular}{lcc}
\toprule
\textbf{Hyperparameter} & \textbf{Score Model} & \textbf{Query Model} \\
\midrule
Epochs & 1 & 4 \\
Batch Size & 16 & 16 \\
Learning Rate & $1 \times 10^{-5}$ & $1 \times 10^{-5}$ \\
Warmup Step Ratio & 0.05 & 0.05 \\
\bottomrule
\end{tabular}
\caption{Training hyperparameters for the score and query models.}
\label{tab:hyperparams}
\end{table}

\subsubsection{Qwen3-embedding training}
For training the Qwen3-Embedding model, all training samples are used only once, i.e., the model is trained for a single epoch. Both training and evaluation are conducted with a maximum input sequence length of 512 tokens. The training is performed using a learning rate of $1 \times 10^{-3}$, with a warm-up ratio of $0.01$ and a fixed warm-up step count of 300.

\subsection{Additional Experimental Results}
\paragraph{Analysis of Queries Redundancy in Synthetic Data.} To evaluate the diversity of generated queries, we compare the frequency distributions of duplicated queries synthesized by two approaches: (1) the baseline supervised fine-tuning (SFT) query model, and (2) our SSRA stage 1 enhanced query model. Table~\ref{tab:query_redundancy} reports the number of unique queries grouped by their frequency of occurrence in a held-out set of 10,000 documents. A lower concentration of repeated queries indicates higher diversity and better generalization in the query generation process.

\begin{table}[h]
\centering

\begin{tabular}{c|r|r}
\toprule
\textbf{Query Frequency} & \textbf{SFT Model} & \textbf{SSRA Stage 1 Model} \\
\midrule
1  & 26,475 & 27,232 \\
2  & 1,308  &   996 \\
3  &   164  &   145 \\
4  &    46  &    38 \\
5  &    17  &    16 \\
6  &    10  &     8 \\
7  &     3  &     1 \\
8  &     2  &     2 \\
9  &     1  &     1 \\
13 &     0  &     1 \\
16 &     1  &     1 \\
26 &     1  &     0 \\
\bottomrule
\end{tabular}
\caption{Frequency distribution of synthetic queries based on their number of repetitions. The SSRA stage 1 model generates fewer duplicated queries compared to SFT, indicating improved query diversity.}
\label{tab:query_redundancy}
\end{table}

\subsection{Dataset}
The \texttt{video\_text} field in our dataset corresponds to a transformed version of the original \texttt{OCR} and \texttt{ASR} signals extracted from the video content, as described earlier. During the transformation process, some \texttt{OCR} or \texttt{ASR} texts lack coherent semantics, resulting in rewriting failures. In such cases, we fall back to a concatenation of the original signals in the format ``\texttt{\{OCR\};\{ASR\}}'', which is then used as the value of the \texttt{video\_text} field in the final dataset.


\end{document}